# Acquiring Receptive Morphology:
# A Connectionist Model

**Michael Gasser**
Computer Science and Linguistics Departments
Indiana University

**Abstract**

This paper describes a modular connectionist model of the acquisition of receptive inflectional morphology. The model takes inputs in the form of phones one at a time and outputs the associated roots and inflections. Simulations using artificial language stimuli demonstrate the capacity of the model to learn suffixation, prefixation, infixation, circumfixation, mutation, template, and deletion rules. Separate network modules responsible for syllables enable to the network to learn simple reduplication rules as well. The model also embodies constraints against association-line crossing.

## Introduction

For many natural languages, a major problem for a language learner, whether human or machine, is the system of bound morphology of the language, which may carry much of the functional load of the grammar. While the acquisition of morphology has sometimes been seen as the problem of learning how to transform one linguistic form into another form, e.g., by [8] and [10], from the learner's perspective, the problem is one of learning how forms map onto meanings. Most work which has viewed the acquisition of morphology in this way, e.g., [1], has taken the perspective of production. But a human language learner almost certainly learns to understand polymorphemic words before learning to produce them, and production may need to build on perception [6]. Thus it seems reasonable to begin with a model of the acquisition of receptive morphology.

In this paper, I will deal with that component of receptive morphology which takes sequences of phones, each expressed as a vector of phonetic features, and identifies them as particular morphemes. This process ignores the segmentation of words into phone sequences, the morphological structure of words, and the the semantics of morphemes. I will refer to this task as root and inflection *identification*. It is assumed that children learn to identify roots and inflections through the presentation of paired forms and sets of morpheme meanings. They show evidence of generalization when they are able to identify the root and inflection of a novel combination of familiar morphemes.

At a minimum, a model of the acquisition of this capacity should succeed on the full range of morphological rule types attested in the world's languages, it should embody known constraints on what sorts of rules are possible in human language, and it should bear a relationship to the *production* of morphologically complex words. This paper describes a psychologically motivated connectionist model (Modular Connectionist Network for the Acquisition of Morphology, MCNAM) which shows evidence of acquiring all of the basic rule types and which also experiences relative difficulty learning rules which seem not to be possible. In another paper [4], I show how the representations that develop during the learning of root and inflection identification can support word production. Although still tentative in several respects, MCNAM appears to be the first computational model of the acquisition of receptive morphology to apply to this diversity of morphological rules. In contrast to symbolic models of language acquisition, it succeeds without built-in symbolic distinctions, for example, the distinction between stem and affix.

The paper is organized as follows. I first provide a brief overview of the categories of morphological rules found in the world's languages. I then present the model and discuss simulations which demonstrate that it generalizes for most kinds of morphological rules. Next, focusing on template morphology, I show how the network implements the analogue of autosegments and how the model embodies one constraint on the sorts of rules that can be learned. Finally, I discuss augmentation of the model with a hierarchical structure reflecting the hierarchy of metrical phonology; this addition is necessary for the acquisition of the most challenging type of morphological rule, reduplication.

## Categories of Morphological Processes

For the sake of convenience, I will be discussing morphology in terms of the conventional notions of **roots**, **inflections**, and **rules**. However, a human language learner does not have direct access to the root for a given form, so the problem of learning morphology cannot be one of discovering how to add to or modify a root. And it is not clear whether there is anything like a symbolic morphological rule in the brain of a language learner.

The following kinds of inflectional or derivational morphological rules are attested in the world's languages: *affixation*, by which a grammatical morpheme is added to a root (or stem), either before (*prefixation*), after (*suffixation*), both before and after (*circumfixation*), or within (*infixation*); *mutation*, by which one or more root segments themselves are modified; *template* rules, by which a word can be described as a combination of a root and a template specifying how segments are to be intercalated between the root segments; *deletion*, by which one or more segments are deleted; *reduplication*, by which a copy, or a systematically and altered copy, of some portion of the root is added to it. Examples of each rule type are included in the description of the stimuli used in the simulations.

## The Model

The approach to language acquisition exemplified in this paper differs from traditional symbolic approaches in that the focus is on specifying the sort of *mechanism* which has the capacity to learn some aspect of language, rather than the *knowledge* which this seems to require. Given the basic problem of what it means to learn receptive morphology, the goal is to begin with a very simple architecture and augment it as necessary. In this paper, I first describe a version of the model which is modular with respect to the identification of root and inflections. The advantages of this version over the simpler model in which these tasks are shared by the same hidden layer is described in a separate paper [5]. Later I discuss a version of the model which incorporates modularity at the level of the syllable and metrical foot; this is required to learn reduplication.

The model described here is connectionist. There are several reasons why one might want to investigate language acquisition from the perspective of connectionism. For the purposes of this paper, the most important is the hope that a connectionist network, or a device making use of a related statistical approach to learning, may have the capacity to learn a task such as word recognition without pre-wired symbolic knowledge. That is, such a model would make do without pre-existing concepts such as **root** and **affix** or distinctions such as regular vs. irregular morphology. If successful, this model would provide a simpler account of the acquisition of morphology than one which begins with symbolic knowledge and constraints.

Words takes place in time, and a psychologically plausible account of word recognition must take this fact into account. Words are often recognized long before they finish; hearers seem to be continuously comparing the contents of a linguistic short-term memory with the phonological representations in their mental lexicons [7]. Thus the task at hand requires a short-term memory of some sort. Of the various ways of representing short-term memory in connectionist networks [9], the most flexible approach makes use of recurrent connections on hidden units. This has the effect of turning the hidden layer into a short-term memory which is not bounded by a fixed limit on the length of the period it can store. The model to be described here is one of the simpler possible networks of this type, a version of the **simple recurrent network** due to [2].

The Version 1 network is shown in Figure 1 Each box represents a layer of connectionist processing units and each arrow a complete set of weighted connections between two layers. The network operates as follows. A sequence of phones is presented to the input layer one at a time. That is, each tick of the network's clock represents the presentation of a single phone. Each phone unit represents a phonetic feature, and each word consists of a sequence of phones preceded by a boundary "phone" consisting of 0.0 activations.

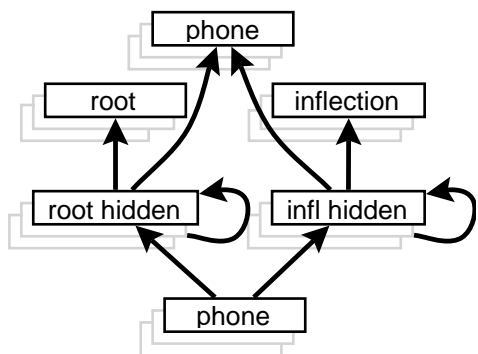

Figure 1: MCNAM: Version 1

An input phone pattern sends activation to the network's hidden layers. Each hidden layer also receives activation from the pattern that appeared there on the previous time step. Thus each hidden unit is joined by a time-delay connection to each other hidden unit within its layer. It is the two previous hidden-layer patterns which represent the system's short-term memory of the phonological context. At the beginning of each word sequence, the hidden layers are reinitialized to a pattern consisting of 0.0 activations.

Finally the output units are activated by the hidden layers. There are at least three output layers. One represents simply a copy of the current input phone. Training the network to auto-associate its current input aids in learning the root and inflection identification task because it forces the network to learn to distinguish the individual phones at the hidden layers, a prerequisite to using the short-term memory effectively. The second layer of output units represents the root "meaning". For each root there is a single output unit. Thus while there is no real semantics, the association between the input phone sequence and the "meaning" is an arbitrary one. The remaining groups of output units represent the inflection "meaning"; one group is shown in the figure. There is a layer of units for each separate inflectional category (e.g., tense and aspect) and a unit for each separate inflection within its layer. One of the hidden layers connects to the root output layer, the other to the inflection output layers.

For each input phone, the network receives a target consisting of the correct phone, root, and inflection outputs for the current word. The phone target is identical to the input phone. The root and inflection targets, which are constant throughout the presentation of a word, are the patterns associated with the root and inflection for the input word.

The network is trained using the backpropagation learning algorithm [11], which adjusts the weights on the network's connections in such a way as to minimize the error, that is, the difference between the network's outputs and the targets.

For each morphological rule, a separate network is trained on a subset of the possible combinations of root and inflection. At various points during training, the network is tested on unfamiliar words, that is, novel combinations of roots and inflections. The performance of the network is the percentage of the test roots and inflections for which its output is correct at the end of each word sequence. An output is considered "correct" if it is closer to the correct root (or inflection) than to any other. The network is evaluated at the end of the word because in general it may need to wait that long to have enough information to identify both root and inflection.

## Experiments

### General Performance of the Model

In all of the experiments reported on here, the stimuli presented to the network consisted of words in an artificial language. The phoneme inventory of the language was made up 19 phones (24 for the mutation rule, which nasalizes vowels). For each morphological rule, there were 30 roots, 15 each of CVC and CVCVC patterns of phones. Each word consisted of either two or three morphemes, a root and one or two inflections (referred to as "tense" and "aspect" for convenience). Examples of each rule, using the root *vibun*: (1) suffix: present–*vibuni*, past–*vibuna*; (2) prefix: present–*ivibun*, past–*avibun*; (3) infix: present–*vikbun*, past–*vinbun*; (4) circumfix: present–*ivibuni*, past–*avibuna*; (5) mutation: present–*vibun*, past–*vibũn*; (6) deletion: present–*vibun*, past–*vibu*; (7) template: present–*vaban*, past–*vbaan*; (8) two-suffix: present perfect–*vibunak*, present progressive–*vibunas*, past perfect–*vibunik*, past progressive–*vibunis*; (9) two-prefix: present perfect–*kavibun*, present progressive–*kivibun*, past perfect–*savibun*, past progressive–*sivibun*; (10) prefix-suffix: present perfect–*avibune*, present progressive–*avibunu*, past perfect–*ovibune*, past progressive–*ovibunu*. No irregular forms were included.

For each morphological rule there were either 60 (30 roots × 2 tense inflections) or 120 (30 roots × 2 tense inflections × 2 aspect inflections) different words. From these 2/3 were selected randomly as training words, and the remaining 1/3 were set aside as test words. For each rule, ten separate networks with different random initial weights were trained and tested. Training for the tense-only rules proceeded for 150 epochs (repetitions of all training patterns); training for the tense-aspect rules lasted 100 epochs. Following training the performance of the network on the test patterns was assessed.

Figure 2. shows the mean performance of the network on the test patterns for each rule following training. Note that chance performance for the roots was .033 and for the inflections .5 since there were 30 roots and 2 inflections in each category. For all tasks, including both root and inflection identification the network performs well above chance. Performance is far from perfect for some of the rule types, but further improvement is possible with optimization of the learning parameters.

Interestingly, template rules, which are problematic for some symbolic approaches to morphology processing and acquisition, are among the easiest for the network. Thus it is informative to investigate further how the network solved this task. For the particular template rule, the two forms of each root shared the same initial and final consonant. This tended to make root identification relatively easy. With respect to inflections, the pattern is more like infixation than prefixation or suffixation because all of the segments relevant to the tense, that is, the /a/s, are between the first and last segment. But inflection identifation for the template is considerably higher than for infixation, probably because of the redundancy: the present tense is characterized by an *a* in second position and a consonant in third position, the past tense by a consonant in second position and an *a* in third position.

To gain a better understanding of the way in which the network solves a template morphology task, a further experiment was conducted. In this experiment, each root consisted of a sequence of three consonants from the set {*p, b, m, t, d, s, n, k, g*}. There were three tense morphemes, each characterized by a particular template. The present template was $C_1aC_2aC_3a$, the past template $aC_1C_2aaC_3$, and the future template $aC_1aC_2C_3a$. Thus the three forms for the root *pmn* were *pamana*, *apmaan*, and *apamna*. The network learns to recognize the tense templates very quickly; generalization is over 90% following only 25 epochs of training. This task is relatively easy since the vowels appear in the same sequential positions for each tense. More interesting is the performance of the root identification part of the network, which must learn to recognize the commonality among sequences of the same consonants even though, for any pair of forms for a given root, only one of the three consonants appears in the same position. Performance reaches 72% on the test words following 150 epochs.

To better visualize the problem, it helps to examine what happens in hidden-layer space for the root layer as a word is processed. This 15-dimensional space is impossible to observe directly, but we can get an idea of the most significant movements through this space through the use of principal component analysis, a technique which is by now a familiar way of analyzing the behavior of recurrent networks [3, 9]. Given a set of data vectors, principal component analysis yields a set of orthogonal vectors, or components, which are ranked in terms of how much of the variance in the data they account for.

Principal components for the root identification hidden layer vectors were extracted for a single network following 150 repetitions of the template training patterns. The paths through the space defined by the first two components of the root identification hidden layer as the three forms of the root *pds* are presented to the network are shown in Figure 3. Points marked in the same way represent the same root consonant.[1] What we see is that, as the root hidden layer processes the word, it passes through roughly similar regions in hidden-layer space as it encounters the consonants of the root, independent of their sequential position. In a sense these regions correspond to the autosegments of autosegmental phonological and morphological analyses.

---

[1] Only two points appear for the first root consonant because the first two segments of the past and future forms of a given root are the same.

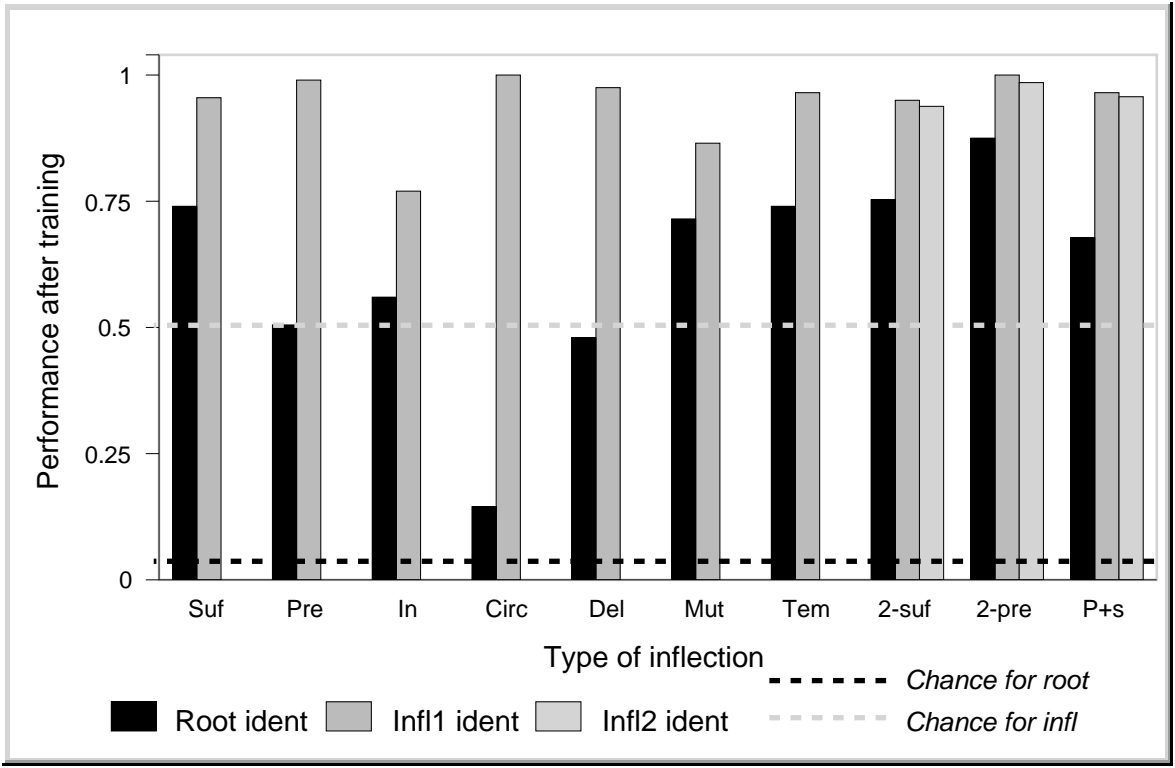

Figure 2: Performance on Test Words Following Training

## Constraints on Morphological Processes

In the previous sections, I have described how modular simple recurrent networks have the capacity to learn to recognize morphologically complex words resulting from a variety of morphological processes. But is this approach too powerful? Can these networks learn rules of types that people cannot? While it is not completely clear what rules people can and cannot learn, some evidence in this direction comes from examining large numbers of languages. One possible constraint on morphological rules comes from autosegmental analyses: the association lines that join one tier to another should not cross. Another way of stating the constraint is to say that the relative position of two segments within a morpheme remains the same in the different forms of the word.

Can a recognition network learn a rule which violates this constraint as readily as a comparable one which does not? To test this, separate networks were trained to learn the following two template morphology rules, involving three forms: (1) present: $C_1aC_2aC_3a$, past: $aC_1C_2aaC_3$, future: $aC_1aC_2C_3a$ (2) present: $C_1aC_2C_3aa$, past: $aC_1C_2aC_3a$, future: $aC_1aC_3aC_2$. Both rules produce the three forms of each root using the three root consonants and sequences of three $a$'s. In each case each of the three consonants appears in the same position in two of the three forms. The second rule differs from the first in that the order of the three consonants is not constant; the second and third consonant of the present and past forms reverse their relative positions in the future form. In the terms of a linguistic analysis, the root consonants would appear in one order in the underlying representation of the root (preserved in the present and past forms) but in the reverse order in the future form. The underlying order is preserved in all three forms for the first rule. I will refer to the first rule as the "favored" one, the second as the "disfavored" one.

In the experiments testing the ease with which these two rules were learned, a set of thirty roots was again generated randomly. Each root consisted of three consonants limited to the set $\{p, b, m, t, d, n, k, g\}$. As before, the networks were trained on 2/3 of the possible combinations of root and inflection (60 words in all) and tested on the remaining third (30 words). Separate networks were trained on the two rules. Mean results for 10 different networks for each rule are shown in Figure 4. While the disfavored rule is learned to some extent, there is a clear advantage for the favored over the disfavored rule with respect to generalization for root identification. Since the inflection is easily recognized by the pattern of consonants and vowels, the order of the second and third root consonants is irrelevant to inflection identification. Root identification, on the other hand, depends crucially on the sequence of consonants. With the first rule, in fact, it is possible to completely ignore the CV templates and pay attention only to the root consonants in identifying the root. With the second rule, however, the only way to be sure which root is intended is to keep track of which sequences occur with which templates. With the two possible roots *ftn* and *fnt*, for example, there would be no way of knowing which root appeared in a form not encountered during training unless the combination of sequence and tense had somehow been attended to during training. In this case, the future of one root has the same sequence of consonants as the present and past of the other. Thus, to the extent that roots overlap with one another, root identification with the disfavored rule presents a harder task to a network. Given the relatively small set of consonants in these experiments, there is considerable overlap among the roots, and this is reflected in the poor generalization for the disfavored rule. Thus for this word recognition network, a rule which apparently could not occur in human language is somewhat more difficult than a comparable one which could.

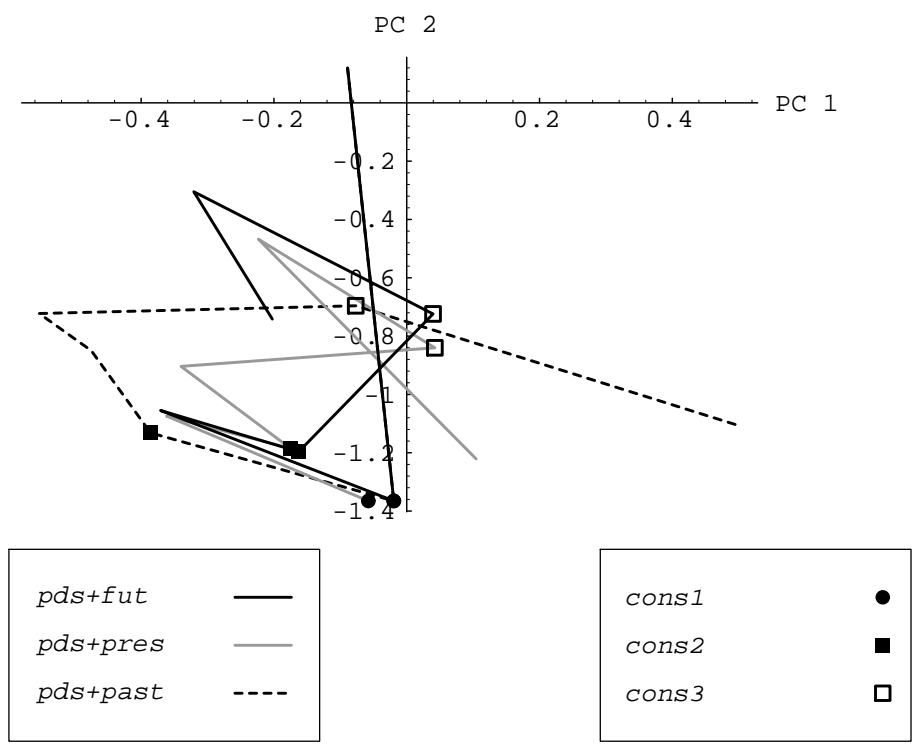

Figure 3: Template Rule, Root Hidden Layer, Principal Components 1 and 2, *padasa, apdaas, apadsa*

## Reduplication

We have yet to deal with reduplication. The parsing of an unfamiliar word involving reduplication apparently requires the ability to notice the similarity between the relevant portions of the word. For the networks we have considered so far, recognition of reduplication would seem to be a difficult, if not an impossible, task. Consider the case in which a network has just heard the sequence *tamkam*. At this point we would expect a human listener to be aware that the two syllables rhymed, that is, that they had the same vowel and final consonant (rime). But at the point following the second *m*, the network does not have direct access to representations for the two subsequences to be compared. If it has been trained to identify sequences like *tamkam*, it will at this point have a representation of the entire sequence in its contextual short-term memory. However, this representation will not distinguish the two syllables, so it is hard to see how they might be compared.

To test whether Version 1 of the model could handle reduplication, networks were trained to perform inflection identification only. The stimuli consisted of two-syllable words, where the initial consonant (the onset) of each syllable came from the set $\{p, b, f, v, m, t, d, s, z, n, k, g, x, \gamma, \eta/\}$, the vowel from the set $\{i, e, u, o, a\}$; and the final consonant, when there was one, from the set $\{n, s\}$. Separate networks were trained to turn on their single output unit when the onsets of the two syllables were the same and when the rimes were the same. The training set consisted of 200 words. In each case, half of the sequences satisfied the reduplication criterion. Results of the two experiments are shown in Figure 5 by the lines marked "Seq". Clearly these networks failed to learn this relatively simple reduplication task. While these experiments do not prove conclusively that a recurrent network, presented with words one segment at a time, cannot learning reduplication, it is obvious that this is a difficult task for these networks.

In a sequential network, input sequences are realized as movements through state space. It appears, however, that recognition of reduplication requires the explicit comparison of *static* representations of the subsequences in question, e.g., for syllables in the case of syllable reduplication. If a simple recurrent network is trained to identify, that is, to distinguish, the syllables in a language, then the pattern appearing on the hidden layer following the presentation of a syllable must encode all of the segments in the syllable. It is, in effect, a summary of the sequence that is the syllable.

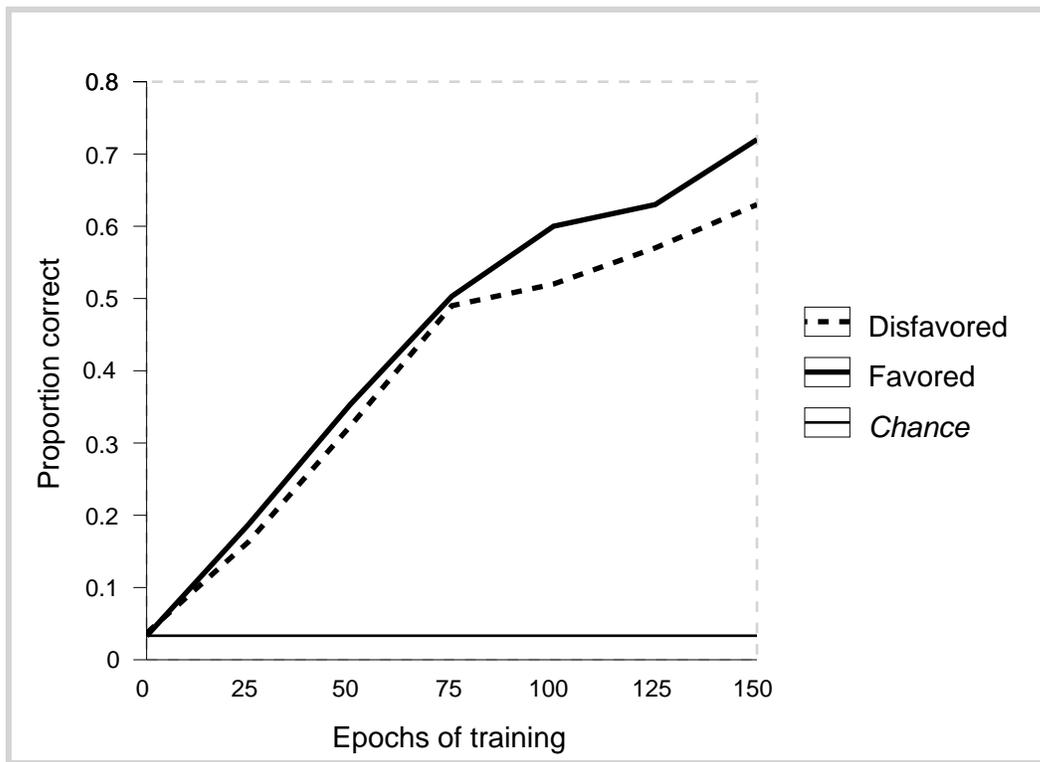

Figure 4: Template Rules, Favored and Disfavored, Root Identification

It is a simple matter to train a network to distinguish all possible syllables in a language. We treat the syllables as separate words in a network like the ones we have been dealing with, but with no inflection module. A network of this type was trained to recognize all 165 possible syllables in the same artificial language used in the experiment with the sequential network. When presented to the network, each syllable sequence was followed by a boundary segment.

The hidden-layer pattern appearing at the end of each syllable-plus-boundary sequence was then treated as a static representation of the syllable sequence for a second task. Previous work [4] has shown that these representations embody the structure of the input sequences in ways which permit generalizations. In this case, the sort of generalization which interests us concerns the recognition of similarities between syllables with the same onsets or rimes. Pairs of these syllable representations, encoding the same syllables as those used to train the sequential network in the previous experiment, were used as inputs to two simple feedforward networks, one trained to respond if its two input syllables had the same onset, the other trained to respond if the two inputs had the same rime, that is, the same rules trained in the previous experiment. Again the training set consisted of 200 pairs of syllables, the test set of 50 pairs in each case. Results of these experiments are shown in Figure 5 by the lines labeled "FF". Although performance is far from perfect, it is clear that these networks have made the appropriate generalization. This means that the syllable representations encode the structure of the syllables in a form which enables the relevant comparisons to be made.

What I have said so far about reduplication, however, falls far short of an adequate account. First, there is the problem of how the network is to make use of static syllable representations in recognizing reduplication. That is, how is access to be maintained to the representation for the syllable which occurred two or more time steps back? For syllable representations to be compared directly, a portion of the network needs to run, in a sense, in syllable time. That is, rather than individual segments, the inputs to the relevant portion of the network need to be entire syllable representations. Combining this with the segment-level inputs that we have made use of in previous experiments gives a hierarchical architecture like that shown in Figure 6. In this network, word recognition, which takes place at the output level, can take as its input both segment and syllable sequences. The segment portion of the network, appearing on the left in the figure, is identical to what we have seen thus far. (Hidden-layer modularity is omitted from the figure to simplify it.) The syllable portion, on the right, runs on a different "clock" from the segment portion. In the segment portion activation is passed forward and error backward each time a new segment is presented to the network. In the syllable portion this happens each time a new syllable appears. (The different update clock is indicated by the dashed arrows in the figure.) Just as the segment subnetwork begins with context-free segment representations, the syllable subnetwork takes as inputs context-free syllables. This is achieved by replacing the context (that is, the recurrent input to the SYLLABLE layer) by a boundary pattern at the beginning of each new syllable.

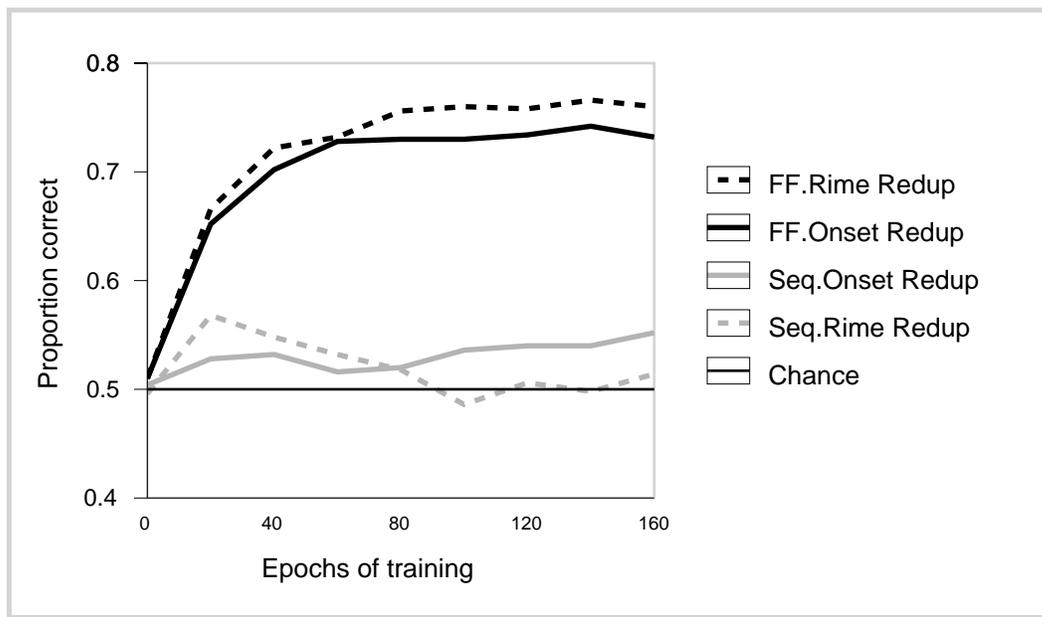

Figure 5: Reduplication Rules, Sequential and Feedforward Networks Trained with Distributed Syllables

There remains the question of how the network is to know when one syllable ends and another begins. Unfortunately this interesting topic is beyond the scope of this project.

## Conclusions

Can connectionist networks which are more than uninteresting implementations of symbolic models learn to generalize about morphological rules of different types? Much remains to be done before this question can be answered, but, for receptive morphology at least, the tentative answer is yes. In place of built-in *knowledge*, e.g, linguistic notions such as affix and tier and constraints such as the prohibition against association line crossing, we have processing and learning algorithms and particular architectural features, e.g., recurrent connections on the hidden layer and modular hidden layers. Some of the linguistic notions may prove unnecessary altogether. For example, there is no place or state in the current model which corresponds to the notion **affix**. Others may be realized very differently from the way in which they are envisioned in conventional models. An autosegment, for example, corresponds roughly to a region in hidden-layer space in MCNAM. But this is a region which took on this significance only in response to the set of phone sequences and morphological targets which the network was trained on.

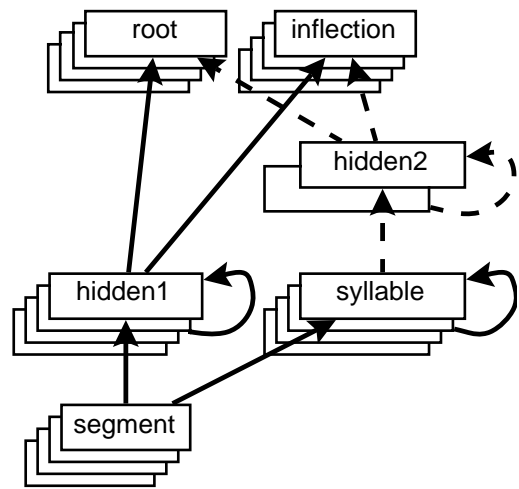

Figure 6: MCNAM: Version 2

Language is a complex phenomenon. Connectionists have sometimes been guilty of imagining naively that simple, uniform networks would handle the whole spectrum of linguistic phenomena. The tack adopted in this project has been to start simple and augment the model when this is called for. MCNAM in its present form is almost certain to fail as a general model of morphology acquisition and processing, but these early results indicate that it is on the right track. In any case, the model yields many detailed predictions concerning the difficulty of particular morphological rules for particular phonological systems, so an obvious next step is psycholinguistic experiments to test the model.